\pdfoutput=1
\documentclass[]{fairmeta}
\usepackage{amsmath,longtable,tabularx}
\usepackage{algorithm,algpseudocode}
\usepackage{tikz}
\usepackage{enumitem}
\usetikzlibrary{arrows.meta,fit}
\title{Engram: Multi-Stage Memory for Agentic Tuning of Heterogeneous Models}
\author{Huanran Li}
\author{Nhat X.T. Le}
\author{Kevin De Angeli}
\author{Yao Xuan}
\author{Alireza Vahdatpour}
\author{Djordje Gligorijevic}
\affiliation{Meta Platforms, Inc.}
\abstract{
Large language model (LLM) agents can now tune neural networks, yet
without cross-session memory they spend costly trials rediscovering known lessons. Reuse is difficult because production models differ in architecture, training setup and inference budget. We present Engram, a memory system that links past trial outcomes with the hypotheses and rationale behind them and assesses whether this evidence applies to the current model. An online agent curates evidence at the decision points where it is needed:
Initial Guidance supplies validated lessons at the start,
Session Retrieval supplies related past sessions before each batch of hypotheses is generated,
and Trial Candidate Review checks each trial candidate against matching past trials. Between sessions, an offline agent distills validated lessons, making the memory self-improving. On two ads models, Engram reaches the 32-trial result of a no-memory control with 47\% fewer trials and also ends with a better model on average.
}
\date{November 2026}
\correspondence{Huanran Li at \email{huanranli@meta.com}}
\hypersetup{pdftitle={Engram: Multi-Stage Memory for Agentic Tuning of Heterogeneous Models},pdfauthor={Huanran Li, Nhat X.T. Le, Kevin De Angeli, Yao Xuan, Alireza Vahdatpour, Djordje Gligorijevic}}
\begin{document}
\maketitle
\section{Introduction}
\label{sec:introduction}
Tuning deep neural networks has long relied on human expertise: engineers choose experiments from experience and the literature and then inspect the results. Automated methods such as Bayesian optimization~\citep{snoek2012practical} reduce this effort but search a fixed, predominantly numeric space that must be specified in advance. Large language model (LLM) agents remove this restriction by reasoning directly over the model and its configuration. Given the current baseline configuration, an objective and a compute budget, an agent reads the model's code and logs, formulates a hypothesis with its rationale, runs a training trial, and uses the outcome to select its next experiment~\citep{react,opro}. Such agents have proven effective: AIDE outperforms H2O AutoML on average across 16 tabular Kaggle tasks~\citep{aide}, and agents for recommendation systems report production launches and an eightfold reduction in training time without loss in business metrics~\citep{selfevolvingrecsys}.

\begin{figure}[!h]
\centering
\includegraphics[width=0.6\linewidth]{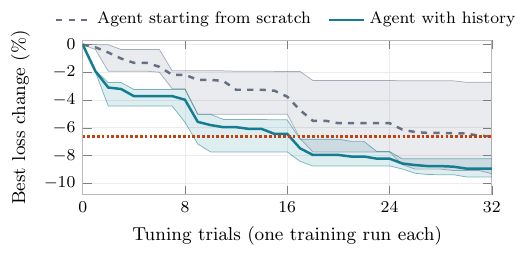}
\caption{Tuning an ads model: best normalized log loss on unseen data (mean and range over five sessions). With history, the agent needs 47\% fewer trials to match its 32-trial result from scratch.}
\label{fig:ne_progress}
\end{figure}

As agents emerge as a tool for model tuning, it is inevitable that they will tune many heterogeneous models, and each of them repeatedly. For example, large advertising platforms maintain an extensive ecosystem of models that serve different product surfaces, ranking stages and advertiser objectives~\citep{liu2026design}. Production models are regularly retrained on fresh data, extended with new features and redesigned around new architectures, and each change calls for a new tuning session. However, reusing earlier sessions is difficult because production models differ in targets, features and architecture, in training setup (data, run length and optimizer schedule), and in the time span of their training data. From our experience in tuning production models, we identify five challenges for agent memory across heterogeneous models. \textbf{(C1) Past gains depend on the specific model and training setup:} the effect of a change such as a larger embedding dimension varies with the model, its baseline and its training setup, so a gain on one model may not recur on another. \textbf{(C2) Applicable changes depend on the inference budget:} a change that helped a model with a generous FLOPs budget may be inapplicable under a tight one. \textbf{(C3) Outcomes alone do not explain a result:} a record of configurations and outcomes rarely reveals the assumptions under which a result should hold, so it cannot tell a later agent whether the result transfers. \textbf{(C4) The relevance of past sessions changes during the current tuning session:} once the current model's outcomes show which changes work, the most useful past sessions are those that gained from the same changes, rather than those most similar at launch or most recent. \textbf{(C5) Misapplied evidence can degrade the search:} past results from models that differ too much from the current one can steer the agent toward worse configurations.
Engram addresses this problem by jointly interpreting past trial outcomes and their rationale against the current model and budget, then delivering applicable evidence at the decision stage where it can guide the search (Figure~\ref{fig:ne_progress}).

\paragraph{Related work.}
Prior work on reusing experience in LLM agents falls into three lines. \emph{LLM agents for model tuning} reuse earlier solutions and outcomes within a task~\citep{aide}, provide a score-sorted journal of past configurations~\citep{selfevolvingrecsys}, or carry distilled heuristics across datasets~\citep{virtualmle}, but none judges whether a result applies to another model or budget (C1, C2, C5). We adopt a journal of past sessions sorted by their performance gain as a baseline to measure what such judgments add. \emph{Experience memory} systems supply distilled insights and similar trajectories~\citep{expel}, rerank and rewrite retrieved experience~\citep{reme}, or learn which memories are useful~\citep{memrl} and when to stop retrieval~\citep{memora}, but perform neither baseline-relative comparison nor budget checks. \emph{Transfer methods for Bayesian optimization} warm-start the search from similar past datasets~\citep{feurer2015init} or weight past tasks by learned similarity~\citep{perrone2018}, but transfer numerical results without the rationale that explains whether a result should transfer (C3).

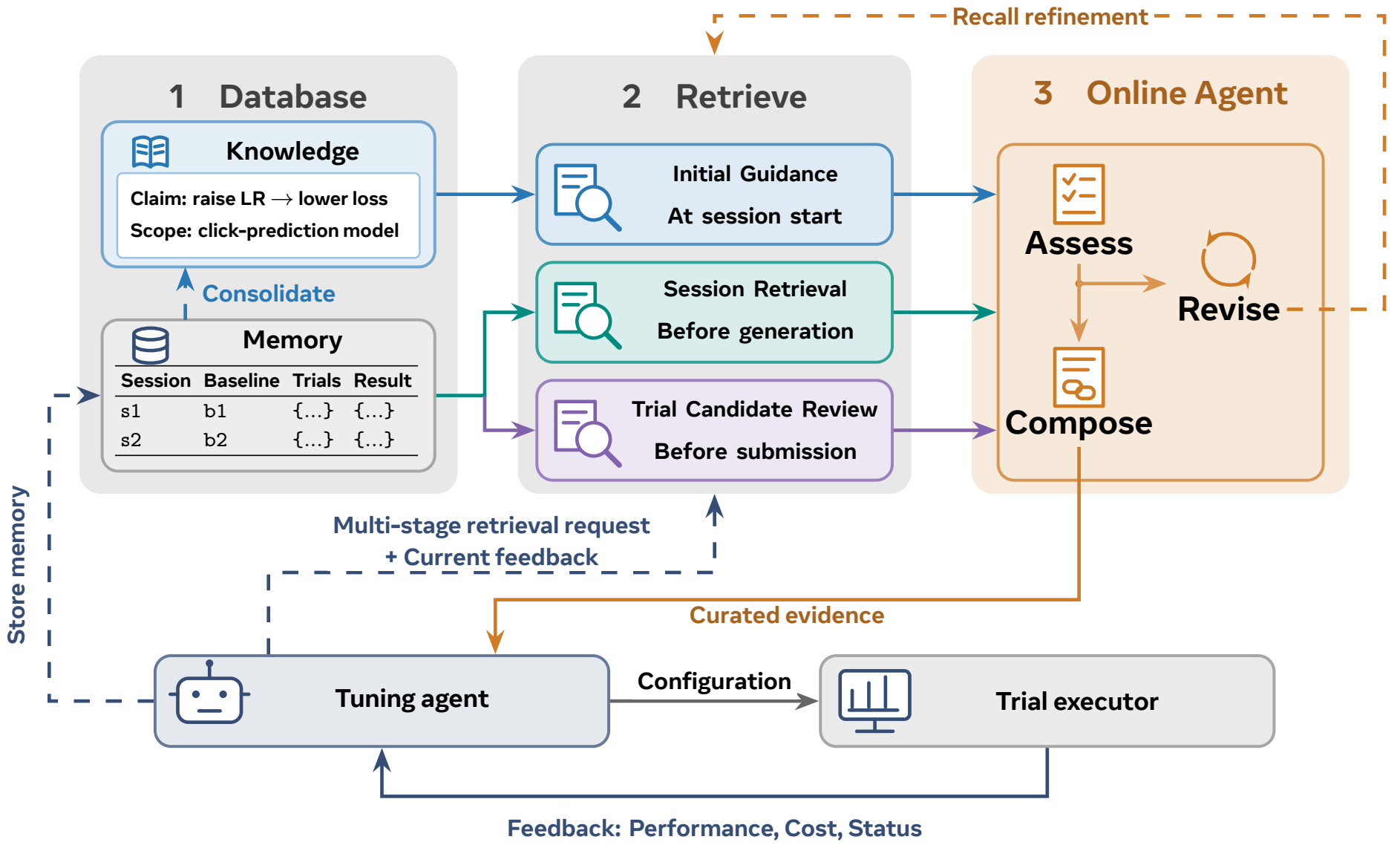
\begin{figure}[!htb]
\centering
\resizebox{0.8\linewidth}{!}{\begingroup
\def\engMemoryTable{{\sffamily\scriptsize
\setlength{\tabcolsep}{2pt}\renewcommand{\arraystretch}{1.18}%
 \begin{tabular}{llll}
 \hline
 \textbf{Session} & \textbf{Baseline} & \textbf{Trials} & \textbf{Result}\\
 \hline
 \texttt{s1} & \texttt{b1} & \texttt{\{...\}} & \texttt{\{...\}}\\
 \texttt{s2} & \texttt{b2} & \texttt{\{...\}} & \texttt{\{...\}}\\
 \hline
 \end{tabular}
}}
\setbox0=\hbox{\engMemoryTable}
\edef\engTableWidth{\the\wd0}
\pgfmathsetmacro{\engInsetScale}{2.03cm/\engTableWidth}
\def\engInsetText#1{\scalebox{\engInsetScale}{{\sffamily\scriptsize #1}}}
\begin{tikzpicture}[
 font=\sffamily\scriptsize,
 every node/.append style={execute at begin node=\hyphenpenalty10000},
 card/.style={draw=black!35,rounded corners=3pt,line width=.55pt,align=center,inner sep=3pt},
 flow/.style={-{Stealth[length=1.6mm]},draw=black!60,line width=.65pt},
 control/.style={flow,dashed},
 label/.style={font=\sffamily\tiny,fill=white,inner sep=1pt},
 stage/.style={font=\sffamily\scriptsize\bfseries,text=black!75}
]
\def\engCardTitle#1{\scalebox{.9}{{\fontsize{6}{7}\selectfont\bfseries #1}}}
\def\engCardSubtitle#1{\scalebox{.8}{{\fontsize{6}{7}\selectfont #1}}}
\def\engStageTitle#1{\scalebox{.78}{{\fontsize{6}{7}\selectfont\bfseries #1}}}
\definecolor{engBlue}{HTML}{2878B5}
\definecolor{engTeal}{HTML}{008B80}
\definecolor{engPurple}{HTML}{8261AF}
\definecolor{engOrange}{HTML}{D17B24}
\definecolor{engInk}{HTML}{354D75}
\tikzset{
 iconline/.style={line width=.65pt,line cap=round,line join=round},
 pics/book/.style={code={
  \draw[iconline,fill=white] (-.2,-.15) -- (-.2,.17) .. controls (-.1,.21) and (-.04,.2) .. (0,.16) .. controls (.07,.2) and (.13,.21) .. (.2,.17) -- (.2,-.15) .. controls (.1,-.11) and (.04,-.12) .. (0,-.16) .. controls (-.07,-.12) and (-.13,-.11) .. cycle;
  \draw[iconline] (0,-.16)--(0,.16);
  \draw[iconline] (-.15,.1)--(-.05,.08) (.05,.08)--(.15,.1) (-.15,.02)--(-.05,0) (.05,0)--(.15,.02);
 }},
 pics/database/.style={code={
  \draw[iconline,fill=white] (-.2,.14)--(-.2,-.14) arc[start angle=180,end angle=360,x radius=.2,y radius=.07]--(.2,.14);
  \draw[iconline,fill=white] (0,.14) ellipse[x radius=.2,y radius=.07];
  \draw[iconline] (-.2,.01) arc[start angle=180,end angle=360,x radius=.2,y radius=.07];
 }},
 pics/search/.style={code={
  \draw[iconline,fill=white] (-.2,-.13) rectangle (.06,.19);
  \draw[iconline] (-.15,.12)--(0,.12) (-.15,.05)--(-.05,.05);
  \draw[iconline,fill=white] (.05,-.06) circle[radius=.11];
  \draw[iconline,line width=1pt] (.13,-.14)--(.22,-.23);
 }},
 pics/shield/.style={code={
  \draw[iconline,fill=white] (0,.22)--(.19,.15)--(.16,-.07).. controls (.12,-.16) and (.04,-.2)..(0,-.23).. controls (-.04,-.2) and (-.12,-.16)..(-.16,-.07)--(-.19,.15)--cycle;
  \draw[iconline,line width=1pt] (-.1,0)--(-.02,-.08)--(.11,.08);
 }},
 pics/checklist/.style={code={
  \draw[iconline,fill=white] (-.16,-.19) rectangle (.16,.19);
  \draw[iconline] (-.1,.1)--(-.07,.07)--(-.03,.13) (.01,.1)--(.1,.1) (-.1,-.01)--(-.07,-.04)--(-.03,.02) (.01,-.01)--(.1,-.01) (-.1,-.12)--(.1,-.12);
 }},
 pics/refresh/.style={code={
  \draw[iconline,-{Stealth[length=1.1mm]}] (.16,-.06) arc[start angle=-20,end angle=150,radius=.17];
  \draw[iconline,-{Stealth[length=1.1mm]}] (-.16,.06) arc[start angle=160,end angle=330,radius=.17];
 }},
 pics/advice/.style={code={
  \draw[iconline,fill=white] (-.16,-.19) rectangle (.16,.19);
  \draw[iconline] (-.1,.12)--(.1,.12) (-.1,.05)--(.06,.05);
  \draw[iconline,rounded corners=1pt] (-.1,-.1) rectangle (.015,-.025) (-.015,-.14) rectangle (.1,-.065);
 }},
 pics/robot/.style={code={
  \draw[iconline,fill=white,rounded corners=2pt] (-.21,-.14) rectangle (.21,.15);
  \draw[iconline] (0,.15)--(0,.23);
  \fill (0,.25) circle[radius=.025];
  \fill (-.09,.045) circle[radius=.025]; \fill (.09,.045) circle[radius=.025];
  \draw[iconline] (-.07,-.065)--(.07,-.065) (-.21,.045)--(-.26,.045) (.21,.045)--(.26,.045);
 }},
 pics/monitor/.style={code={
  \draw[iconline,fill=white,rounded corners=1pt] (-.23,-.12) rectangle (.23,.2);
  \draw[iconline] (0,-.12)--(0,-.2) (-.12,-.2)--(.12,-.2);
  \draw[iconline] (-.15,-.04)--(-.15,.07) (-.04,-.04)--(-.04,.12) (.07,-.04)--(.07,.16);
  \draw[iconline] (-.17,-.06)--(.15,-.06);
 }}
}
\fill[black!9,rounded corners=4pt] (.05,1.82) rectangle (2.55,4.72);
\fill[black!9,rounded corners=4pt] (2.95,1.82) rectangle (5.55,4.72);
\fill[engOrange!16,rounded corners=4pt] (5.95,1.82) rectangle (8.45,4.72);
\node[stage] at (1.3,4.45) {1\quad Database};
\node[stage] at (4.25,4.45) {2\quad Retrieve};
\node[stage,text=engOrange!80!black] at (7.2,4.45) {3\quad Online Agent};

\node[card,fill=engBlue!12,draw=engBlue!65,minimum width=2.2cm,minimum height=.96cm] (guides) at (1.3,3.8) {};
\pic[color=engBlue,scale=.55] at (.52,4.07) {book};
\node at (1.46,4.07) {\engCardTitle{Knowledge}};
\path[draw=engBlue!45,fill=white,line width=.35pt,rounded corners=1pt] (.30,3.38) rectangle (2.30,3.94);
\node[anchor=west,inner sep=0pt] at (.38,3.77) {\engInsetText{Claim: raise LR $\to$ lower loss}};
\node[anchor=west,inner sep=0pt] at (.38,3.55) {\engInsetText{Scope: click-prediction model}};
\node[card,fill=black!7,minimum width=2.2cm,minimum height=1.0cm] (records) at (1.3,2.47) {};
\pic[color=engInk,scale=.55] at (.52,2.80) {database};
\node at (1.46,2.82) {\engCardTitle{Memory}};
\node[inner sep=0pt] at (1.3,2.37) {\resizebox{2.03cm}{!}{\engMemoryTable}};
\draw[control,draw=engBlue] (.75,2.97) -- (.75,3.32);
\node[label,anchor=west,fill=none,text=engBlue] at (.82,3.145) {Consolidate};

\node[card,fill=engBlue!15,draw=engBlue!80,minimum width=2.35cm,minimum height=.66cm] (t1) at (4.25,3.8) {};
\pic[color=engBlue] at (3.4,3.8) {search};
\node[align=center,text width=1.8cm] at (4.52,3.8) {\engStageTitle{Initial Guidance}\\\engCardSubtitle{At session start}};
\node[card,fill=engTeal!15,draw=engTeal!80,minimum width=2.35cm,minimum height=.66cm] (t2) at (4.25,3.02) {};
\pic[color=engTeal] at (3.4,3.02) {search};
\node[align=center,text width=1.8cm] at (4.52,3.02) {\engStageTitle{Session Retrieval}\\\engCardSubtitle{Before generation}};
\node[card,fill=engPurple!15,draw=engPurple!80,minimum width=2.35cm,minimum height=.66cm] (t3) at (4.25,2.24) {};
\pic[color=engPurple] at (3.4,2.24) {search};
\node[align=center,text width=1.8cm] at (4.52,2.24) {\engStageTitle{Trial Candidate Review}\\\engCardSubtitle{Before submission}};
\draw[flow,draw=engBlue] (guides.east) -- (t1.west);
\draw[flow,draw=engTeal] (records.east) -- (2.73,2.47) |- (t2.west);
\draw[flow,draw=engPurple] (2.73,2.47) |- (t3.west);

\node[card,fill=engOrange!17,draw=engOrange!80,minimum width=2.15cm,minimum height=2.22cm] (curator) at (7.2,3.02) {};
\pic[color=engOrange] at (6.66,3.80) {checklist};
\node at (6.66,3.48) {Assess};
\pic[color=engOrange] at (7.65,3.37) {refresh};
\node[inner sep=1pt] (revise) at (7.65,3.04) {Revise};
\pic[color=engOrange] at (6.66,2.59) {advice};
\node[inner sep=1pt] (compose) at (6.66,2.27) {Compose};
\coordinate (decision) at (6.66,3.21);
\fill[engOrange] (decision) circle[radius=.025];
\draw[flow,draw=engOrange!80] (6.66,3.34) -- (decision) -- (7.24,3.21);
\draw[flow,draw=engOrange!80] (decision) -- (6.66,2.82);
\draw[flow,draw=engBlue] (t1.east) -- (curator.west |- t1.east);
\draw[flow,draw=engTeal] (t2.east) -- (curator.west |- t2.east);
\draw[flow,draw=engPurple] (t3.east) -- (curator.west |- t3.east);
\draw[control,draw=engOrange] (revise.east) -- (8.68,3.04) -- (8.68,4.98) -- (4.25,4.98) -- (4.25,4.72);
\node[label,text=engOrange!80!black] at (6.47,4.98) {Recall refinement};

\node[card,fill=engInk!13,draw=engInk!75,minimum width=3.0cm,minimum height=.60cm] (agent) at (2.05,.45) {};
\pic[color=engInk] at (.91,.45) {robot};
\node at (2.25,.45) {\engCardTitle{Tuning agent}};
\node[card,fill=black!8,minimum width=3.0cm,minimum height=.60cm] (executor) at (6.45,.45) {};
\pic[color=engInk] at (5.31,.45) {monitor};
\node at (6.65,.45) {\engCardTitle{Trial executor}};
\draw[flow,draw=engOrange] (compose.south) -- (6.66,1.12) -- node[label,below,text=engOrange!80!black] {Curated evidence} (2.8,1.12) -- (agent.north -| 2.8,0);
\draw[control,draw=engInk] (agent.north -| 1.3,0) -- (1.3,1.30) -- node[label,above,align=center,text=engInk] {Multi-stage retrieval request\\+ Current feedback} (4.25,1.30) -- (4.25,1.82);
\draw[control,draw=engInk] (agent.west) -- (-.15,.45) -- (-.15,2.47) -- (records.west);
\node[label,rotate=90,text=engInk] at (-.35,1.35) {Store memory};
\draw[flow] (agent.east) -- node[label,above] {Configuration} (executor.west);
\draw[flow,draw=engInk] (executor.south) -- ++(0,-.32) -| (agent.south);
\node[label,text=engInk] at (4.25,-.40) {Feedback: Performance, Cost, Status};
\end{tikzpicture}
\endgroup}
\caption{Online Agent: multi-stage retrieval and curation of evidence for the tuning agent. Insets are schematic.}
\label{fig:overview}
\end{figure}

Figure~\ref{fig:overview} gives an overview of Engram.
Our contributions are as follows:
\begin{itemize}[leftmargin=1em,itemsep=0pt,parsep=0pt,topsep=2pt]
\item \textbf{Joint analysis of trial outcomes and rationale:} a relational store links the outcome of every past trial with its hypothesis and rationale, and an LLM-based online agent uses both to judge whether a recalled result is well supported and applicable to the current model and budget (C1, C3).
\item \textbf{Multi-stage retrieval:} Initial Guidance (IG) supplies validated lessons whose scope matches the current model at launch, Session Retrieval (SR) retrieves related past sessions and re-ranks them by current feedback before each batch, and Trial Candidate Review (TCR) checks each trial candidate against matching past trials and the budget (C2, C4, C5).
\item \textbf{Self-improving memory through knowledge consolidation:} between sessions, an offline agent derives lessons from completed sessions, validates them on later sessions and applies them in subsequent ones.
\item \textbf{Preliminary evidence:} Engram outperforms a no-memory control early in the search on two ads models and a journal sorted by performance on the model with a poor history.
\end{itemize}

\section{Engram Design}
\label{sec:method}
\subsection{Problem Setting}
\label{sec:background}
A tuning session starts from a baseline configuration $x_0$ in a fixed evaluation context and may submit at most $K$ trials, each evaluating one configuration $x_k$ of search parameters such as learning rates and layer widths. Each trial measures a quality metric $q(x)$, for which lower is better, and an inference cost $c(x)$; a trial is feasible if its cost does not exceed $\rho$ times that of the baseline. The agent chooses each trial using only the outcomes of trials completed so far, and the session returns
\begin{equation}
\label{eq:objective}
x^\star=\operatorname*{arg\,min}_{x\in\{x_0,\dots,x_K\}}\ q(x)\quad\text{s.t.}\quad c(x)\le\rho\,c(x_0).
\end{equation}
In our setting, $q$ is the normalized entropy (NE), the log loss divided by the entropy of the average label rate~\citep{he2014practical}, and $c$ is the inference FLOPs; we use these names below. The tuning agent runs an observe--reason--act loop~\citep{react} with the LLM as an optimizer~\citep{opro}: it reads a tuning guide of general advice and the outcomes of the current session, writes each hypothesis with its rationale, and submits trials. Engram complements this feedback with evidence from past sessions, which the agent would otherwise not see.
\subsection{Memory, Knowledge and Consolidation}
\label{sec:stores}
Engram maintains two stores that separate raw evidence from validated lessons. \emph{Memory} holds the raw evidence, which a daily backfill ingests from every concluded session. For each trial, it records the configuration changes, NE, FLOPs and status together with the tuning agent's hypothesis, rationale and analysis.  The rationale is written before a trial runs and explains why the hypothesis should improve NE, whereas the analysis interprets the measured outcome afterwards. \emph{Knowledge} in turn holds the lessons that offline consolidation derives from \emph{Memory} and validates. Each lesson specifies a parameter, a range of starting values and a change to apply from that range, which together define a move. It also states a claim about the move's effect on NE and a scope that defines the models to which the claim applies, such as models of a given architecture or trained on a given amount of data.

\paragraph{Offline consolidation}
\label{sec:consolidation}
This process distills evidence across all models into global lessons and stores each lesson in \emph{Knowledge} only after validating it. In this way, the offline agent turns \emph{Memory} into \emph{Knowledge}, which makes Engram a \textit{self-improving memory} whose guidance strengthens as it tunes more models. Code first mines candidate lessons from pairs of trials within a session, measuring how each move changed NE relative to the configuration it modified. Single-parameter and joint changes are mined separately, and lessons from joint changes are stated only as correlations because their effect cannot be attributed to one parameter. The offline agent then reviews each candidate with its supporting trials and recorded rationale, keeps it, narrows its scope or drops it, and relates it to the default advice in the tuning guide. A candidate is published only if it (1) predicts outcomes more accurately than the default tuning guide, both on the full history and under temporal cross-validation, (2) draws on multiple distinct models (five by default), and (3) is endorsed by multiple independent LLM reviews (two by default).

\begin{algorithm}[!htb]
\caption{Engram in one tuning session}
\label{alg:engram}
\begin{algorithmic}[1]
\Require baseline configuration $x_0$, trial budget $K$, FLOPs cap $\rho$
\State Compose the \textit{Initial Guidance (IG)} brief
\For{each batch of trials within the budget $K$}
\State Run \textit{Session Retrieval (SR)} on the outcomes observed so far
\Repeat
\State The tuning agent proposes trials or revises flagged ones
\State Run \textit{Trial Candidate Review (TCR)} on the proposed trials
\Until{no trial is flagged or the review limit is reached}
\State Execute the trials and record their outcomes
\EndFor
\State Store the session in \emph{Memory} for offline consolidation
\State \Return $x^\star$, the best feasible configuration
\end{algorithmic}
\end{algorithm}

\subsection{\texorpdfstring{Multi-Stage}{Multi-Stage} Retrieval}
\label{sec:stages}
Algorithm~\ref{alg:engram} shows when each stage acts: Initial Guidance runs once when a session starts, after which the session proceeds in rounds until its trial budget is exhausted. In each round, Session Retrieval retrieves past sessions given the outcomes so far, the tuning agent proposes a batch of trials, and Trial Candidate Review checks it before submission. At each stage, an LLM-based online agent receives the retrieved evidence with the current baseline and outcomes and operates in a loop: it assesses whether the evidence applies to the current model, revises the retrieval when evidence is insufficient, and composes the text that the tuning agent reads. We next describe how each stage retrieves and curates its evidence.

\paragraph{Initial Guidance (IG)}
When a session starts, Initial Guidance informs the tuning agent which validated lessons can be applied to its model. A lesson from the latest consolidation run applies if (1) its scope covers the current model, (2) the model exposes the lesson's parameter, and (3) the baseline value of that parameter lies within the move's starting range. The online agent reviews all applicable lessons against the baseline configuration and composes a brief that recommends the moves these lessons support and reports joint-change lessons as correlations rather than recommendations.

\paragraph{Session Retrieval (SR)}
Before the tuning agent generates each batch of hypotheses, Session Retrieval retrieves past sessions that resemble the current one and explains what they found. It ranks each past session $c$ by a weighted metadata match plus a weighted agreement of hypotheses and outcomes:
\begin{equation}
\label{eq:recall}
\mathrm{score}(c)=\sum_i w_i\,\mathrm{Match}_i(c)+\lambda\,\mathrm{Agreement}(c).
\end{equation}
Each $\mathrm{Match}_i(c)\in[0,1]$ compares one metadata field of $c$ with that of the current session, such as the model architecture, the model scale or the explored parameters. The weight $w_i$ sets the importance of each field. $\mathrm{Agreement}(c)$ measures how much of the current session's NE gain so far comes from hypotheses that changed parameters which $c$ also explored and improved. The weight $\lambda$ sets its importance relative to the metadata match. Because Agreement is recomputed after every batch, the ranking evolves with new outcomes. The online agent receives the ranked sessions with their measurements and rationale, checks whether they cover the parameters that can be tuned on the current model, fit the FLOPs budget and remain recent, and may exclude sessions or adjust $w_i$ and $\lambda$ to retrieve a revised set. It then writes advice citing the measurements and rationale of the retained sessions, while the tuning agent keeps the choice of the next hypothesis.

\paragraph{Trial Candidate Review (TCR)}
After the tuning agent proposes a batch of trials, Trial Candidate Review checks each trial candidate against past trials in \emph{Memory} that made the same change. Whereas Session Retrieval reasons about whole sessions, Trial Candidate Review examines the individual trials that bear on a proposed change. The online agent receives each candidate together with past trials that moved the same parameters. It judges whether the candidate is likely to improve NE and stay within the FLOPs cap, and the tuning agent must replace or justify every candidate judged unlikely to improve NE or over the cap.

\begin{table}[!htb]
\caption{Experiment arms and mean best feasible NE change (\%) over five sessions after 8, 16, 24 and 32 trials. Bold: ahead of the control (exact Mann--Whitney U, $p<0.05$, uncorrected).}
\label{tab:preliminary_results}
\centering\footnotesize
\setlength{\tabcolsep}{2pt}
\begin{tabular*}{0.75\linewidth}{@{\extracolsep{\fill}}lrrrrrrrr@{}}
\toprule
 & \multicolumn{4}{c}{Model 1} & \multicolumn{4}{c}{Model 2}\\
\cmidrule(lr){2-5}\cmidrule(l){6-9}
Arm & 8 & 16 & 24 & 32 & 8 & 16 & 24 & 32\\
\midrule
\multicolumn{9}{@{}l}{\textit{Baselines}}\\
Control (no memory) & $-2.17$ & $-3.75$ & $-5.66$ & $-6.61$ & $-0.29$ & $-0.63$ & $-0.85$ & $-1.11$\\ %
NE-sorted journal & $\mathbf{-3.55}$ & $\mathbf{-6.01}$ & $-7.91$ & $-9.09$ & $-0.32$ & $-0.65$ & $-0.96$ & $-1.26$\\ %
\midrule
\multicolumn{9}{@{}l}{\textit{Engram stages}}\\
IG only & $-2.97$ & $-5.88$ & $-7.62$ & $-8.66$ & $\mathbf{-0.48}$ & $-0.78$ & $-0.95$ & $-1.15$\\ %
SR only & $\mathbf{-4.17}$ & $\mathbf{-6.48}$ & $-7.97$ & $-9.01$ & $-0.32$ & $-0.61$ & $-1.11$ & $-1.34$\\ %
IG + SR & $\mathbf{-3.47}$ & $\mathbf{-5.85}$ & $-7.37$ & $-9.03$ & $\mathbf{-0.44}$ & $\mathbf{-0.95}$ & $-1.18$ & $-1.35$\\ %
Full Engram & $\mathbf{-3.97}$ & $\mathbf{-6.43}$ & $\mathbf{-8.22}$ & $-8.93$ & $-0.40$ & $\mathbf{-0.96}$ & $\mathbf{-1.33}$ & $\mathbf{-1.46}$\\ %
\midrule
\multicolumn{9}{@{}l}{\textit{Ablations of full Engram}}\\
No curation & $\mathbf{-3.49}$ & $-5.89$ & $-7.66$ & $-8.63$ & $-0.41$ & $-0.65$ & $-1.10$ & $-1.33$\\ %
No agreement & $\mathbf{-4.04}$ & $-5.95$ & $-7.42$ & $-8.76$ & $\mathbf{-0.48}$ & $-0.89$ & $-1.10$ & $-1.27$\\ %
Distant history only & $\mathbf{-4.79}$ & $\mathbf{-6.40}$ & $-7.59$ & $-8.65$ & \multicolumn{4}{c}{Same as full Engram}\\ %
\bottomrule
\end{tabular*}
\end{table}

\section{Experiments}
\label{sec:evaluation}
We examine whether historical evidence helps an agent find useful tuning hypotheses earlier and which stage supplies the evidence that matters. We call a past model \emph{close} to the current one if both belong to the same production model type, and \emph{distant} otherwise. A model type fixes the ad product, ranking stage, predicted event and attribution window, and hence the traffic it serves; close models still differ in architecture, features and training regime. We study two ads models that differ in their history. Model 1 serves as the main study and has a rich history of 227 sessions, 59 of which come from close models with eight other baselines. Model 2 tests whether the findings generalize when the history is poor: all 299 of its sessions come from distant models, although some share its architecture family. Neither history includes the current evaluation baseline. Table~\ref{tab:preliminary_results} summarizes the arm design.

\subsection{Experimental Setup and Metrics}
\label{sec:setup}
Both models are evaluated with eight arms: the no-memory control, an NE-sorted journal in place of Engram, IG only, SR only, IG with SR, full Engram, and full Engram without online curation or without feedback agreement. Model 1 additionally runs full Engram with distant history only; on Model 2 this arm coincides with full Engram because the entire history is distant. Each arm comprises five sessions that submit trials in synchronized batches of eight under an inference-FLOPs cap of $\rho=1.5$. Each trial trains for a single epoch on about 190M examples for Model 1 and about 144M examples for Model 2, and a batch takes about five and three hours, respectively. Apart from the batch size, the number of epochs and the FLOPs cap, the search space is unrestricted; agents mostly adjust learning rates (LR), warmup schedules, optimizer settings, embedding and layer sizes, and input features. The tuning agent and its subagents use Claude Opus 5.5, and the online agent uses the smaller Claude Sonnet 5.5 to reduce Engram's cost and latency, since each of its calls judges supplied evidence without tool use. Within each session, Initial Guidance runs once, Session Retrieval retrieves up to 50 sessions and revises its retrieval at most twice per batch, and Trial Candidate Review checks each batch once. In practice, tuning sessions on these models run a median of 9--10 trials, and 90\% end within 25, so early gains matter most; we analyze 32 trials to verify that they persist.

\paragraph{Metrics.}
Because each trial trains for a single epoch and every example is scored before the model trains on it, NE is always measured on unseen data. We report NE as a percentage change relative to the baseline, so negative values indicate improvements. We report each arm's mean best feasible NE change after 8, 16, 24 and 32 trials, which reflects both search speed and final quality, with differences in percentage points. We compare two arms with a two-sided exact Mann--Whitney U test on their five session outcomes each. We use a rank test because five sessions cannot support a normality assumption and the outcomes are often bimodal, since sessions either find the decisive change or stall.

\subsection{Speed and Final Quality}
Engram reaches strong configurations earlier on both models, and on Model 2 it also achieves a better final result (Table~\ref{tab:preliminary_results}). On both models, full Engram's mean reaches the control's 32-trial mean with 47\% fewer trials (17 instead of 32). On Model 1, full Engram, SR alone, IG with SR, the NE-sorted journal and the distant-only arm all lead the control after 8 and 16 trials ($p\le0.008$), and full Engram still leads at 24 trials. The control then narrows the gap: three of its sessions independently discover the decisive LR lever (Section~\ref{sec:evidence_reach}) and reach $-8.89\%$ to $-9.30\%$, whereas the other two never do and remain near $-3\%$. 
On Model 2, neither stage suffices alone. IG alone leads the control only at 8 trials and reaches the control's 32-trial mean only at trial 30, whereas SR alone stays level with the control through 16 trials. Full Engram, by contrast, leads the control at 16 and 24 trials ($p=0.008$), still leads at 32 trials ($p=0.032$; $-1.46\%$ against $-1.11\%$), and leads both single stages at 16 and 24 trials ($p=0.008$). 
The NE-sorted journal shows when Engram's applicability judgments matter. It matches full Engram at every checkpoint on Model 1 ($p\ge0.22$), where 59 past sessions come from close models. On Model 2, where the entire history is distant, it never leads the control significantly and trails full Engram at 16, 24 and 32 trials ($p\le0.032$).

\subsection{How Evidence Reaches the Search}
\label{sec:evidence_reach}

The decisive evidence originates from different stages on the two models, which supports multi-stage retrieval. For each decisive hypothesis, Table~\ref{tab:evidence_paths} reports which stage supplied the evidence and how many control and full-Engram sessions test it and when. This retrospective analysis does not isolate the effects of individual stages. On Model 1, Initial Guidance supplies a lesson to triple the learning rate of non-embedding parameters (the dense LR) from its baseline value of 0.04, beyond the tuning guide's range of 0.02--0.08 and into the region where the guide warns of divergence. Every full-Engram session raises the LR to 0.12 or more (first at trials 10--25), whereas only three control sessions do so (at trials 18--29), and the two that never do finish last. With IG alone, however, the lead over the control is not significant at any checkpoint ($p\ge0.056$). Session Retrieval additionally supplies the appropriate embedding size from close-model sessions: every full-Engram session adopts it in its first trial, whereas the control and the distant-only arm find it only at trials 2--7 and spend 21 and 11 trials on larger sizes, respectively. 
On Model 2, the initial LR lesson merely moves the first LR increase to trial 1; the decisive change is instead shortening both the dense and the sparse LR warmups. This schedule appears only in history from distant models: every full-Engram session adopts it at trials 10--15 and cites a single recalled session as its source, whereas three control sessions discover it independently from trial 18 onward and SR alone first makes it only at trials 13--26. 

\begin{table}[!htb]
\caption{Decisive hypotheses within 32 trials: stage, sessions testing it, median first trial ($>$32 if never) as control\,/\,Engram, and mean NE change of trials making it. Bold: Engram ahead.}
\label{tab:evidence_paths}
\centering\footnotesize
\setlength{\tabcolsep}{2pt}
\begin{tabular*}{0.75\linewidth}{@{\extracolsep{\fill}}clcccr@{}}
\toprule
Model & Hypothesis & Stage & Sessions & 1st trial & NE (\%)\\
\midrule
1 & Dense LR $0.04\to{\ge}0.12$ & IG & 3\,/\,\textbf{5} & 29\,/\,\textbf{18} & $-8.29$\\
1 & Embedding size $4\to8$ & SR & 5\,/\,5 & 4\,/\,\textbf{1} & $-5.74$\\
2 & Dense LR $>0.04$ & IG & 5\,/\,5 & 6\,/\,\textbf{1} & $-0.73$\\
2 & Shorter LR warmups & SR & 3\,/\,\textbf{5} & 28\,/\,\textbf{11} & $-1.17$\\
\bottomrule
\end{tabular*}
\end{table}

\section{Conclusion}
Engram links measurements with their rationale, curates evidence at three stages of tuning and validates lessons offline. On two ads models, it reaches the 32-trial result of a no-memory control with 47\% fewer trials, including on a model whose history comes entirely from distant models, and its decisive evidence originated from different stages on each model.
These preliminary findings cover two models with five sessions per arm; broader transfer and the benefit of the self-improving memory across
repeated consolidation remain to be evaluated.
Engram's central design principle is to assess which past evidence applies and deliver it when the tuning
agent needs it, rather than merely supply more history.
\section*{Ethics Statement}
Engram reads only the configurations, metrics and rationale of completed tuning sessions and collects no user data.
\bibliographystyle{plainnat}
\bibliography{references}
\end{document}